\def\year{2019}\relax
\documentclass[letterpaper]{article} 
\usepackage{aaai19}  
\usepackage{times}  
\usepackage{helvet}  
\usepackage{courier}  
\usepackage{url}  
\usepackage{graphicx}  
\usepackage{times}
\usepackage{xcolor}
\usepackage{soul}
\usepackage[utf8]{inputenc}
\usepackage[small]{caption}
\usepackage{amsmath}
\usepackage{amssymb}
\usepackage{graphicx}
\usepackage{booktabs}
\usepackage{multirow}
\usepackage{array}
\usepackage{bbm}
\usepackage{multicol}
\usepackage{blindtext}
\usepackage[labelfont={small,bf},font={small,bf},skip=0pt]{caption}
\usepackage{algorithm}
\usepackage{algorithmic}
\usepackage{tikz}
\usepackage{lipsum}
\usepackage{ntheorem}
\usepackage{makecell}

\title{ A Coarse to Fine Question Answering System based on Reinforcement Learning}

\author{Yu Wang and Hongxia Jin\\ 
Samsung Research America  \\
$\lbrace$yu.wang1, hongxia.jin$\rbrace$@samsung.com}

\begin{document}

\maketitle

\begin{abstract}
In this paper, we present a coarse to fine question answering (CFQA) system based on reinforcement learning which can efficiently processes documents with different lengths by choosing appropriate actions. The system is designed using an actor-critic based deep reinforcement learning model to achieve multi-step question answering. Compared to previous QA models targeting on datasets mainly containing either short or long documents, our multi-step coarse to fine model takes the merits from multiple system modules, which can handle both short and long documents. The system hence obtains a much better accuracy and faster trainings speed compared to the current state-of-the-art models. We test our model on four QA datasets, WIKEREADING, WIKIREADING LONG, CNN and SQuAD, and demonstrate 1.3$\%$-1.7$\%$ accuracy improvements with 1.5x-3.4x training speed-ups in comparison to the baselines using state-of-the-art models.
\end{abstract}
\section{Introduction}
Machine comprehension based question answering (QA) tasks have drawn lots of interests from the natural language understanding research community. During the past several years, lots of progresses have been made on building deep learning based QA models to answer questions from large scale datasets including unstructured documents \cite{hermann2015teaching,hill2015goldilocks,onishi2016did,trischler2016newsqa,nguyen2016ms}. Currently, most of the state-of-the-art models for different QA datasets are based on recurrent neural networks which can process the sequential inputs, and with a (co-)attention structure to deal with the long term interactions between questions and document context  \cite{xiong2016dynamic,chen2016thorough,hermann2015teaching,kadlec2016text}. One disadvantage of these models is that their training and inference speeds are relatively slow due to their recurrent nature, and the other weakness is that they are still not good at dealing with very long documents and model of the models use the truncated documents as their inputs \cite{miller2016key,hewlett2016wikireading}. 

In \cite{choi2017coarse}, it introduces a coarse-to-fine question answering model to select the related sentences in the long document first, then find the answers from the selected sentences, which helps reduce the computational workload. The fast model can be implemented using bag-of-words (BoW) or convolutional neural network and the slow model uses an RNN based model. The model gives decent performance on long documents in WIKIREADING dataset, and shows significant speed-up compared to earlier models. This coarse to fine model, however, still have several disadvantages:

1.The model doesn't perform as good as baseline models on the full WIKIREADING dataset including both short and long documents \cite{hewlett2016wikireading}. One reason is because that the conventional RNN based models perform better on generating answers from the first few sentences of a document, hence can obtain the correct results on short documents more accurately than a coarse-to-fine model . 

2. Many wrong answers are actually from the same sentence containing the correct answer. By using a coarse-to-fine model on sentence level, it may still extract wrong answer by ignoring the correct one in the same sentence.

3. Some of the contextual information in document is still useful and necessary in order to find the correct answers. However, by only selecting the related sentence(s), the system may ignore some important information to make the correct judgment.

On the other hand, compared to the coarse-to-fine model which gives decent results on long documents, more question answering models focus on improving performance on relatively short documents. QANet \cite{wei2018fast} uses the convolutions and self-attentions to boost the training speed and gives one of the state-of-the-art results on the SQuAD1.1 dataset \cite{rajpurkar2016squad}, which contains mainly short documents (average 122 tokens/document). 



Due to these reasons, in this paper, we would like to introduce a novel mult-step coarse to fine question answering (CFQA) system which can handle both long and short documents. This new structure takes the advantages from both coarse-to-fine and QANet structures depending the lengths of documents. Furthermore, a false-positive detection mechanism is also designed to further improve the model performance.

In order to guide the model to learn what is the best action to perform at each step, three types of states are defined as:\\
1. Generate answers directly from the answer generation module (Terminal State).\\
2. Select the related sentences and use them for next round (Non-terminal State)\.\\
3. Remove the inaccurate answer generate from the answer generation module, and use the rest of the document for next round (Non-terminal State).

The first two states are corresponding to the QANet answer generation module and a coarse sentence selector in the coarse-to-fine model. The last state is designed to handle the case where the predicted answer given by the answer generator is a false positive one, which should be removed from current document context to avoid confusion.

At each round of the decision making process, the system performs an action and reaches one out of the three states. If it is a non-terminal state, the system will continue to search the correct answer from the current context, otherwise it will generate the answer from the answer generation module, which is a terminal state. A detailed system description will be given in next section.

In this paper, we propose a DRL based multi-step coarse-to-fine QA (CFQA) system, which mainly gives three contributions:

1. It is the first multi-step QA system which has the capability to decide whether to generate a fine result or a coarse one based on the current context. By using this technique, the model can handle both short and long documents in faster and a more robustness manner. To the best of our knowledge, we are the first one to do so.

2. Besides generating answers from coarse or fine context, the system also learn another ability to reject the false-positive answer given by the answer generator. This self-correctness ability makes the system more "smart".

3. The system achieves significant better QA accuracy compared to our baseline models on four QA datasets, and 1.5x-3.4x training speed-up due to the improved efficiency by reallocating of training data using three actions.
\section{The System and Model Structure}
\begin{figure}
\begin{minipage}{.49\textwidth}
  \centering
  \includegraphics[width=0.9\linewidth]{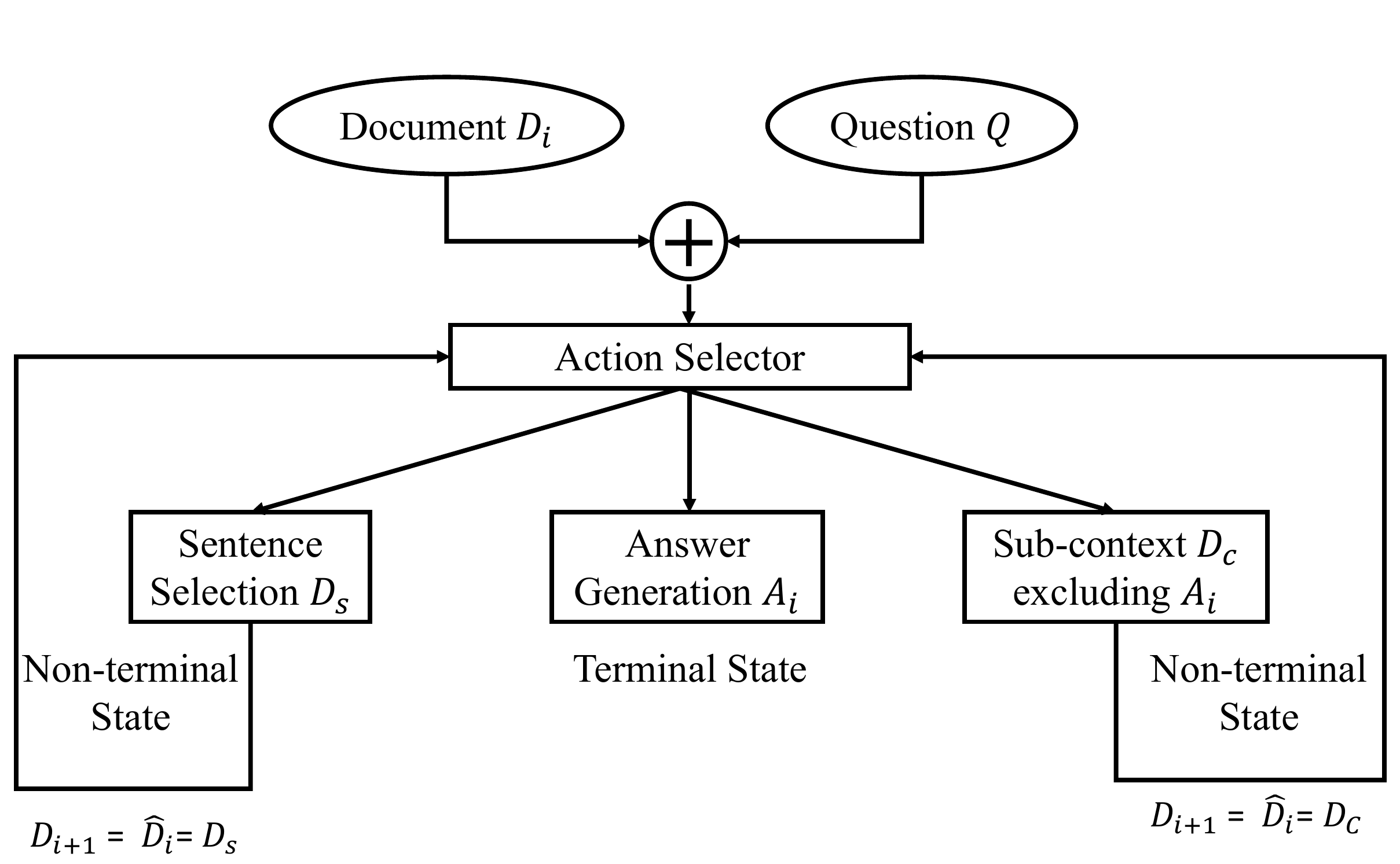}
  \caption{System Diagram}
  \label{fig:systemDiagram}
\end{minipage}
\end{figure}

In this section, the system structure of our multi-step coarse-to-fine QA system is given. The system contains four parts: the first one is a DRL based action selector, the second one is a sentence selection module ($M_1$), the third one is an answer generation module ($M_2$) and the last one is the sub-context generation module ($M_3$). Both the sub-context and sentence selection module are non-terminal states, and the answer generation module is a terminal state. Our system diagram is as given in Figure \ref{fig:systemDiagram}.
\subsection{Action Selector}
The action selector is designed to generate one action to pick one of the three  modules at each time step. In order to model the process of selecting an action, we choose a deep reinforcement learning (DRL) based algorithm to learn it automatically. Formulated as a markov decision process (MDP) , a DRL based model mainly contains four key elements: state $s_t$, action $a_t$, reward $r_t$ and policy $\pi$. In a given a state $s_t$ during a stochastic process, the system is seeking its best action $a_t$ to perform in order to maximize its expected rewards to be obtained, by following some policy $\pi$. The main target of a DRL based model is to seek the best policy $\pi^*$, hence the corresponding action $a^*$, for an agent to perform. There are mainly three types of reinforcement learning algorithms: value-based, policy gradient and actor-critic. We choose the actor-critic based DRL model in order to obtain a relative stable training result for a  large state space as in our system.

The definition of the states $s_t$ and the actions $a_t$ in our system will be given in the section. We will also discuss how to train the action selection module using our selected deep reinforcement learning model, \emph{i.e.} an Actor-Critic based DRL model.
\subsubsection{States ($s_t$)}
The design of the state is as given in Figure \ref{fig:State}, which is the input of the action selection module. It mainly contains two parts: one is the encoded document information, and the other is the question information. The encoded document ($\hat{D}_i$) at step $i$ have three different types of varieties:

\begin{figure}
\begin{minipage}{.49\textwidth}
  \centering
  \includegraphics[width=0.7\linewidth]{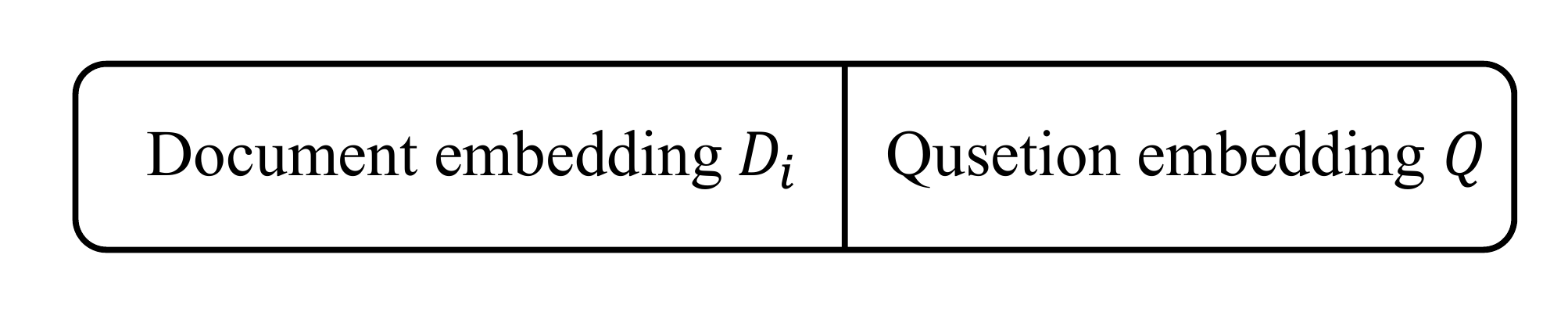}
  \caption{Design of State}
  \label{fig:State}
\end{minipage}
\end{figure}
\underline{Type 1:} It contains the full document information in step $i$, \emph{i.e.} $\hat{D}_i=D_i$. At step 0, this document is the the same as the original given document ($D$), \emph{i.e.} $D_0=D$. 

\underline{Type 2:} It contains the selected sentences information, $\emph{i.e.}$ $\hat{D}_i= D_s=\cup_{i=i}^K s_i$, where $s_i$ is the $i^{th}$ selected sentence from the current document $D_i$, and $K$ is the total number of sentences selected.

\underline{Type 3:} It contains the current document $D_i$ excluding the answer $A_{i}$ predicted by $D_i$, i.e. $\hat{D}_i=D_c=D_i\setminus A_{i}$.

Again, the third type of state's definition is based on the fact that most of the correct answers are within the top $K$ possible answers, if the one with highest probability is not correct one. By defining three possible actions, it gives our system more potentials to find the correct answers both from document level (Type 1) and sentence level (Type 2) by rejecting the incorrect answers (Type 3).

Another important part of a state is the encoded question embedding ($Q$), which doesn't change during the multi-step procedure to guarantee that the system infers the answer from the same given question. 

Both the document and question are encoded using two layers structure. The first layer is an input embedding layer and the second layer is an embedding encoder layer.  The input embedding layer is used to generate each word's representation by concatenating its word-level embedding $x_w$ and character-level embedding $x_c$, \emph{i.e.} $x_i=x_w \oplus x_c\in \mathbb{R}^{d_1+d_2}$, where $d_1$ is the dimension of word-level embedding, $d_2$ is the dimension of the character-level embedding, and $``\oplus"$ stands for the concatenation of two vectors. The word embedding $x_w$ uses the pre-trained GloVe \cite{pennington2014glove} embedding vector where $d_1=300$, and each character of the word is represented by a trainable $d_2=200$ dimension vector. The character level word representation $x_c$ is then generated by taking the maximum value of each row of the character matrix of a word. The word-level vector and character-level vector combine together as the input of the embedding encoder layer.

\begin{figure}
\begin{minipage}{.49\textwidth}
  \centering
  \includegraphics[width=1\linewidth]{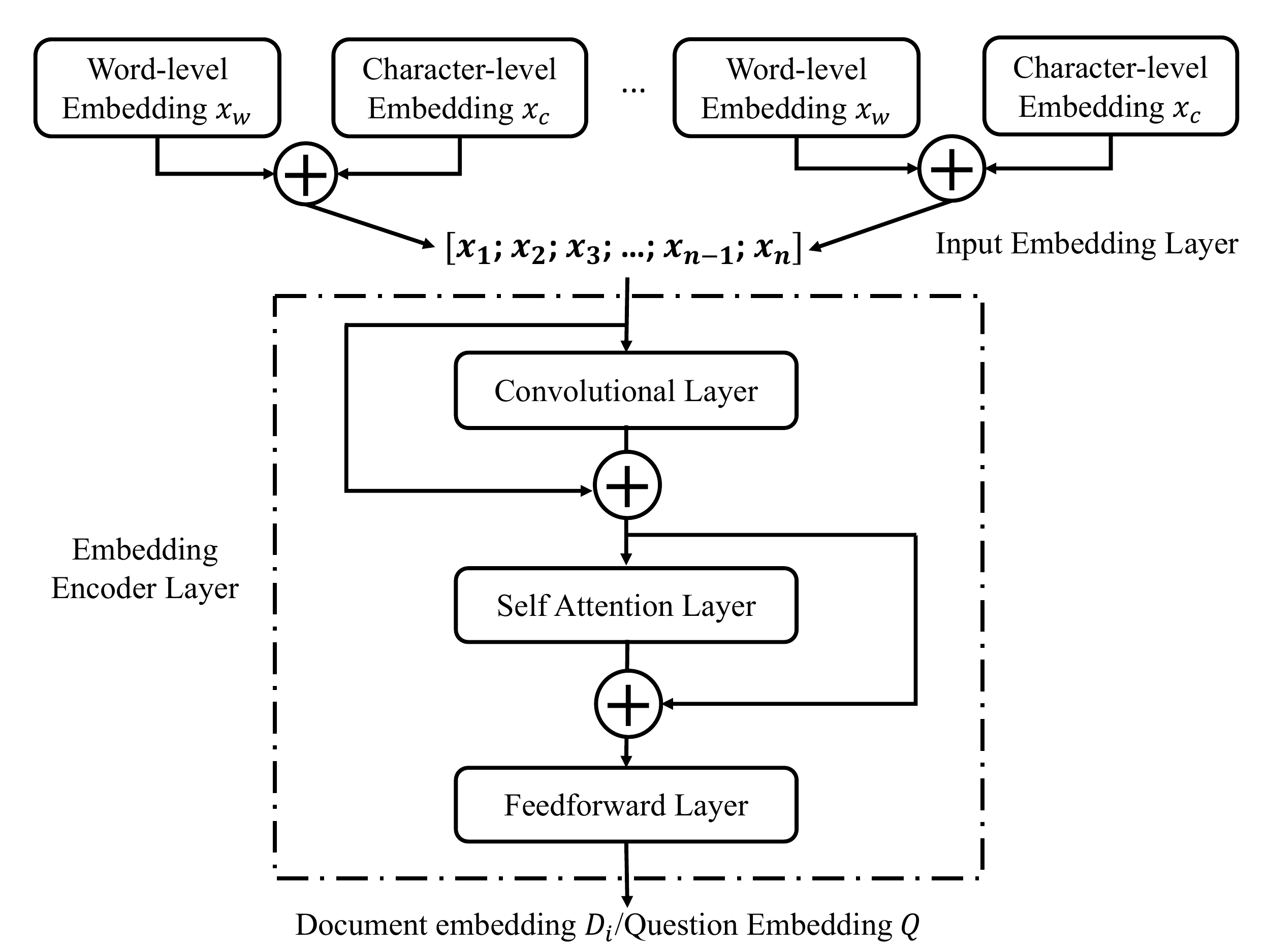}
  \caption{Design of Input Embedding Layer and Embedding Encoder Layer}
  \label{fig:inputEmbedding}
\end{minipage}
\end{figure}

The embedding encoder layer takes the concatenated embeddings of tokens in a document or question as its input $x=[x_1;x_2;\cdots;x_n]$, and pass through several layers as illustrated in Figure \ref{fig:inputEmbedding}. The set-up is similar to that in \cite{wei2018fast} except that here we only use one convolution layer to boost the training speed. The word vector $x$ firstly passed a convolution layer, then passed by a self-attention layer, and finally a feed-forward layer. The convolution layer has a kernel size $k_s$, and the number of filters is $d_f$. The self-attention layer adopt the multi-head attention mechanism as given in \cite{vaswani2017attention}. The output of the last layer is the encoded document representation $\hat{D}_i$ in a state $s_t$ or the question representation $Q$, depending whether the input is a document or the given question.

\subsubsection{Actions ($a_t$)}
As described earlier, three actions are defined in our system to be selected from at each time step $t$. Their definitions are given in Figure \ref{fig:Action}, and the three possible scenarios are as given below:

\begin{figure}
\begin{minipage}{.49\textwidth}
  \centering
  \includegraphics[width=0.9\linewidth]{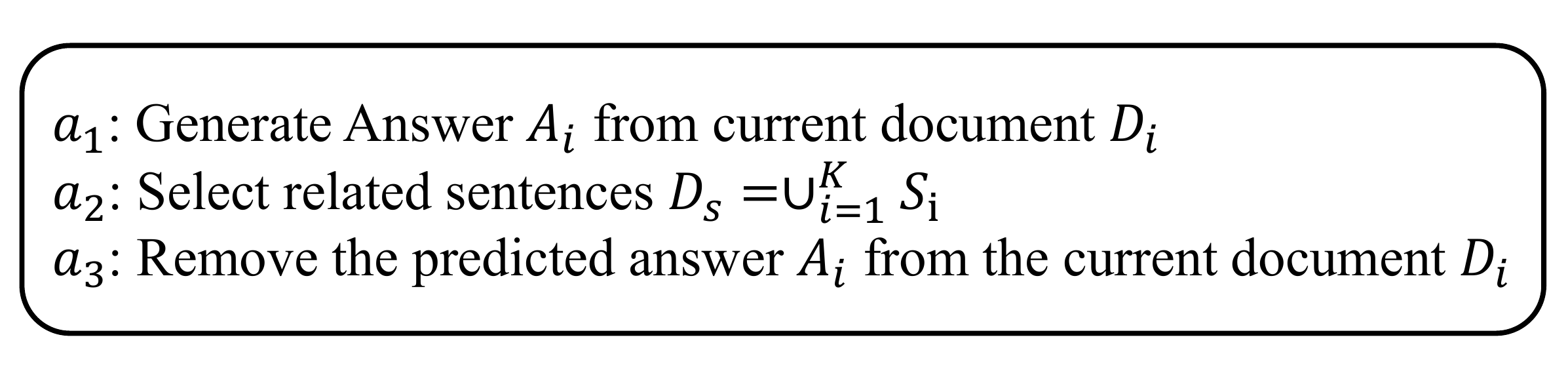}
  \caption{Design of Action}
  \label{fig:Action}
\end{minipage}
\end{figure}

\underline{Action $a_1$:} The first action is to generate the answer directly. The answer generation module will read in the word level embedding for the current context of document $\hat{D}_i$ and question $Q$, then generate two probabilities for each word as the starting/ending point of the answer.  The final answer $ A_{i}$ will be given as the context between the words with highest starting and end probabilities.

\underline{Action $a_2$:} The second action is the sentence selection action. It will select the sentences with highest possibilities containing the correct answers. Assuming $K$ sentences are selected at this step, the new document context will be $\hat{D}_i=[\hat{s}_1;\hat{s}_2;\hat{s}_3;...;\hat{s}_K]$. A detailed description will be given in next section. This action will generate a non-terminal state.

\underline{Action $a_3$:} The last action is to remove the potential answer $A_i$ generated from the current context $\hat{D}_i$, as the system believes that the answer is a false positive answer, it will remove it to avoid confusion.

It is worth noticing that all three actions corresponding to generating the three types of states as described before ($a_1\rightarrow M_2$, $a_2\rightarrow M_1$ and $a_3\rightarrow M_3$). During the training, our system can decide what is the best action to choose based on the expected rewards.
\subsubsection{Rewards ($r_t$)}
Another important elements in reinforcement learning is the reward $r_t$, which can greatly affect the performance and robustness of a DRL model. In our system, different rewards are designed for three types of action-state pairs:

1.If action $a_1$ is selected, \emph{i.e.} to generate the answer directly, the reward is then defined as the F1-score between generate answer $A_i$ and the ground truth $A^*$, \emph{i.e.}:
\begin{equation}
r_1=F1(A_i,A^*)\;\;\;\;\;\;\;\text{if}\;\;\; a=a_1
\end{equation}

2. If the action is chosen to select related sentences ($a_2$), then the reward is defined as whether the selected new context $\hat{D}_i= D_s=\cup_{i=i}^K s_i$ contains the ground truth $A^*$. If it contains the answer, the reward is 1, otherwise it is 0:
\begin{equation}
r_2=
    \begin{cases}
      1 & \text{if $A^*\subseteq D_s=\cup_{i=i}^K s_i$}\\
     0 & \text{otherwise}\\
    \end{cases}
\end{equation}

3. If the action is chosen to remove the potential answer $A_i$ generated from the current context $D_i$, then reward is defined as if the new context $\hat{D}_i=D_c=D_i\setminus A_{i}$ contains the ground truth answer $A^*$. If it contains the answer, the reward is 1, otherwise it is 0, \emph{i.e.}
\begin{equation}
r_3=
    \begin{cases}
      1 & \text{if $A^*\subseteq D_c=D_i\setminus A_{i}$}\\
      0 & \text{otherwise}\\
    \end{cases}
\end{equation}
It is worth noticing that the first reward is assigned to the case when answer is generated directly at the first step.  The second and third rewards are assigned to the states before the last answer generation step if there are multiple steps. The reason is because we will only assign one final reward to a sample to guarantee the convergence of our RL algorithm.
\subsection{Sentence Selection Module $M_1$}
\begin{figure}
\begin{minipage}{.49\textwidth}
  \centering
  \includegraphics[width=0.9\linewidth]{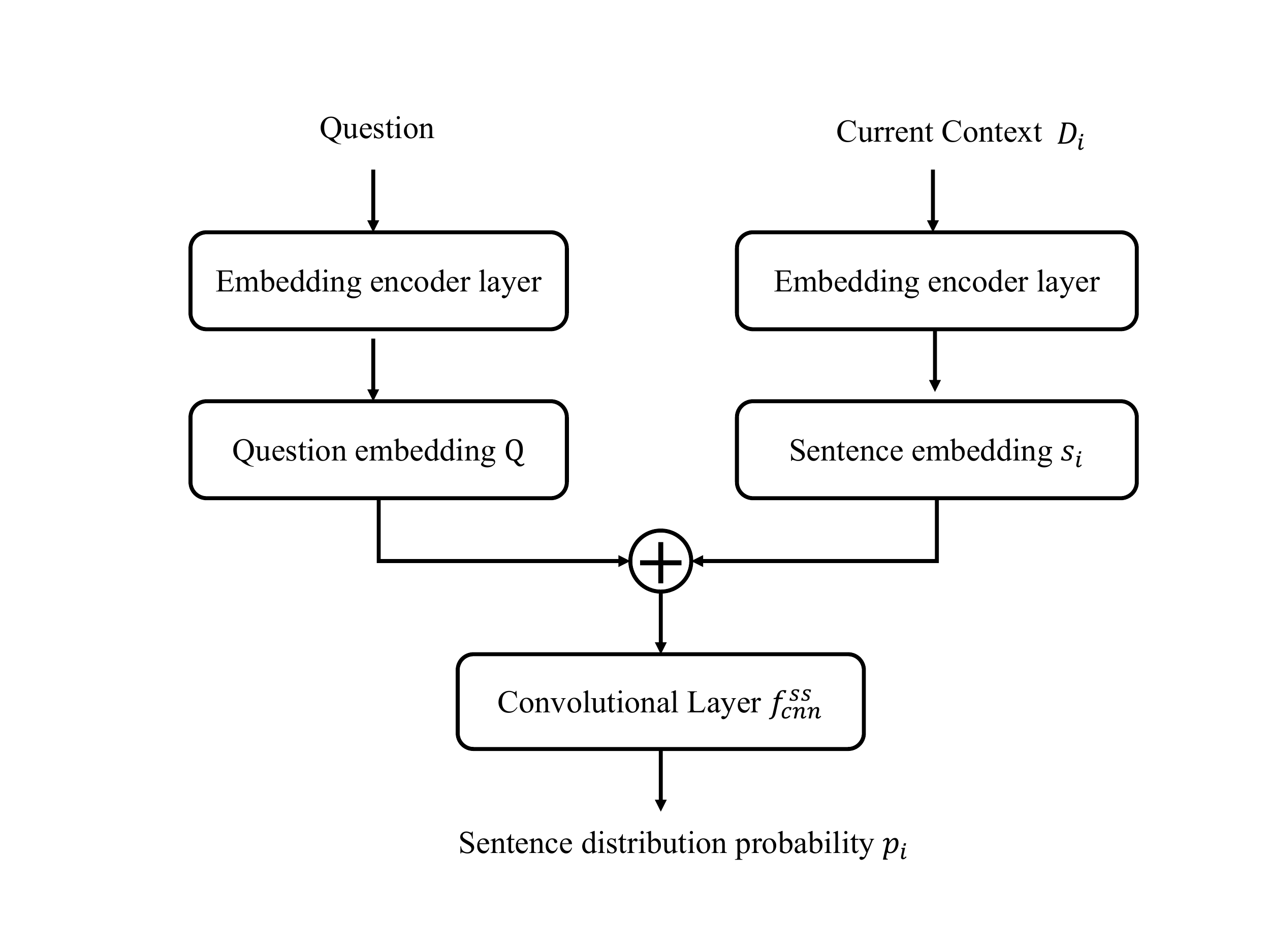}
  \caption{Structure of the Sentence Selector Module $M_1$}
  \label{fig:sentenceSelector}
\end{minipage}
\end{figure}

If the action selector choose action $a_2$, our system will trigger the sentence selection module $M_1$, where multiple answer-related sentences are picked out.
Following recent work on sentence selection \cite{yu2014deep,choi2017coarse,yang2016hierarchical}, a convolutional neural network $f_{cnn}^{ss}$ is used to define a distribution over the sentences $\lbrace s_1,\cdots,s_N \rbrace$, where $N$ is the total number of sentences in current document. Each of the $N$ sentences in current document context together with the question embedding $Q$ will be fed into the convolutional neural network, whose outputs are the probability distributions of all sentences $p(s=s_i|x,d) (i=\lbrace 1,\cdots,N\rbrace)$. To represent it mathematically:
\begin{equation}
p(s=s_i|Q,D_i)=f_{cnn}^{ss}(Q,D_i)
\end{equation}
To explain it in detail, the embeddings of tokens in a question $Q$ are concatenated with those in a sentence $s_i$, then used as the input to $f_{cnn}^{ss}$. The structure is as shown in Figure \ref{fig:sentenceSelector}.
During the training process, the sentences with top $K$ highest probabilities are selected and combined as the document context in next step, \emph{i.e.} $\hat{D}_i= D_s=\cup_{i=i}^K s_i$. The predefined value $K$ is reduced by 1 at each action step, \emph{i.e.} one fewer sentence is selected after one round. The convolutional layer $f_{cnn}^{ss}$ is trained together in the system using the actor-critic's loss functions which are given in next section.
\subsection{Answer Generation Module $M_2$}
\begin{figure}
\begin{minipage}{.49\textwidth}
  \centering
  \includegraphics[width=0.95\linewidth]{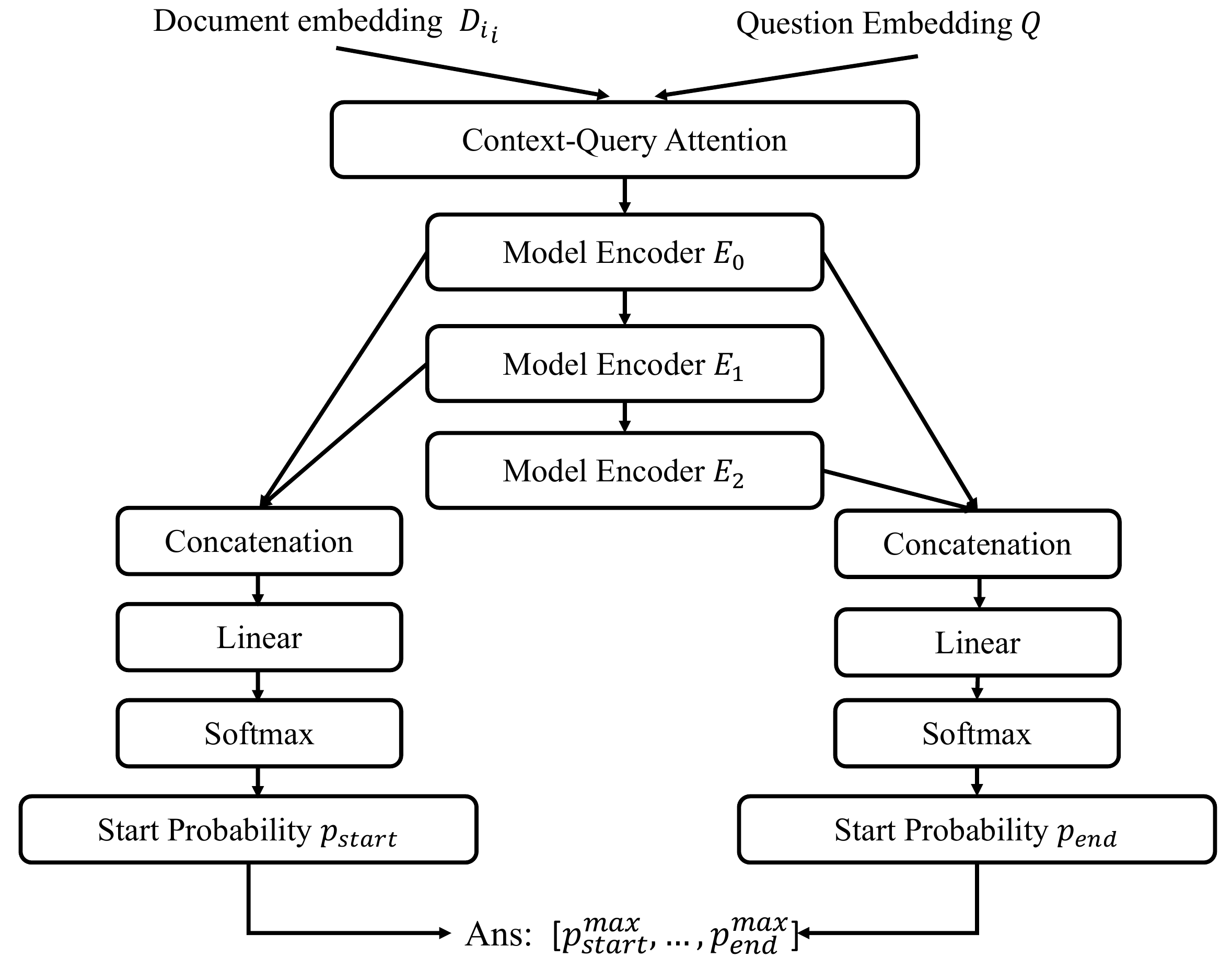}
  \caption{Structure of the Answer Generation Module $M_2$}
  \label{fig:answerGeneration}
\end{minipage}
\end{figure}

If our system decides to answer the question directly based on current context $D_i$ by choosing action $a_1$, the answer generation module will be triggered. Currently, most of the state-of-the-art question answering models use RNN models to encode the document and question and generate the answer. Here, we follow the QANET structure that gives the state-of-the-art result using single model on the SQuAD1.1 dataset \cite{rajpurkar2016squad}. The structure contains a context-query attention layer, several model encoder layers and an output layer. The structure is as shown in Figure \ref{fig:answerGeneration}. Due to the space limitation, we only give a brief overview about the context-query attention layer and model encoder layer, all other details can refer to \cite{wei2018fast}.
\subsubsection{Context-Query Attention Layer}
This is a standard module in many earlier reading comprehension models \cite{weissenborn2017making,chen2017reading,wei2018fast}. Assuming that the encoded context and question embeddings are $D_i$ and $Q$. The context-query attention is computed by using the similarity matrix $\bar{S}$ generated from each pair of context and query words by using their normalized similarities (after applying a softmax layer to each row of their original similarity matrix $S$). The context-to-query attention $A$ can be computed as $A=\bar{S}\cdot Q^T$, and the similarity function is a trilinear function:
\begin{equation}
f(q,d)=W_0[q,d,q\odot d]
\end{equation}
where $q$ is the word token in question $Q$, and $d$ is the word token in current context $D_i$, $\odot$ is the element-wise multiplication and $W_0$ is a trainable variable.

Similarly, by applying the softmax layer to each column of their original similarity matrix $S$, we can obtain a column normalized similarity matrix $\bar{\bar{S}}$. The query to context attention can be computed as $B=\bar{S}\cdot \bar{\bar{S}}^T\cdot D_i^T$.
\subsubsection{Model Encoder Layer}
Here we follow the structure of model encoder layer designed in \cite{seo2016bidirectional}, where the input of the model encoder layer at each position is: $[d,a,d\odot a, d\odot b]$, where $a$ and $b $ are a row of attention matrix $A$ and $B$. Similar to the embedding encoder for document and question as given previously,  the model encoder block for answer generation also contains a convolutional layer, a self-attention layer and a feedforward layer. There are three repetitive encoder blocks which share the same weights between each other. The final outputs of the last layer of model are the probabilities of the starting and ending positions, \emph{i.e.}
\begin{equation}
\begin{split}
p_{start}&=softmax(W_1[E_0;E_1])\\
p_{end}&=softmax(W_2[E_0;E_2])
\end{split}
\end{equation}
The final answer are the extracted tokens between the two words with highest starting probability $p_{start}^{max}$ and ending probability $p_{end}^{max}$. This generated answer is used to define the reward $r_1$, which will be further used to define our loss function.
\subsection{Sub-context Selection Module $M_3$}
Besides the answer generation module (action $a_1$), sentence selection module (action $a_2$), the last module corresponding to the state after conducting action $a_3$ is the sub-context selection module, which is also used to take care of the cases that the answer $A_i$ generated by the current document $D_i$ has a higher possibility as a false positive answer, hence should be eliminated.

Once our action selection module choose action $a_3$, the sub-context selection module will be triggered. The system firstly generate an answer $A_i$ by calling the answer generation module ($M_2$) based on current document context $D_i$ at step $i$, then return a new document context $\hat{D}_i$ by removing the answer $A_i$ and concatenating the sub-contexts before and after $A_i$, \emph{i.e.}  $\hat{D}_i=D_i\setminus A_i$. This new document context is fed back to the action selection module as the document context in next round, that is $D_{i+1}=\hat{D}_i$.
\subsection{Training}
 In this paper, we use the actor-critic based algorithm \cite{konda2000actor,peters2008natural}  to build our DRL model.  Unlike the value based DRL approach like DQN \cite{mnih2015human} or policy based approach like policy gradient \cite{sutton2000policy}, an actor-critic based model has a better performance on continuous state space in terms of convergence and stability. Due to the space limitation and the focus of this paper, we will not spend too much time on explaining the fundamental details of the actor-critic method and only a brief overview is given.
\subsubsection{Actor-Critic}
In actor-critic based RL algorithm, two neural networks are used to model the actor and critic separately. The actor model performs like the policy gradient method by taking state $s_t$ as its input and generating action probabilities based on the policies: $\pi_\theta(a_t|s_t)$. Comparatively, the critic model is similar to a value based approach (like $DQN$): after performing the actor model generated state $a_t$, the system will reach a new state $s_{t+1}$, and the critic model can generate two values (\emph{i.e.}expect rewards) $v_t$ and $v_{t+1}$  by taking $s_t$ and $s_{t+1}$ as inputs. It is worth noticing that there are two loss functions for training actor neural network and critic neural network separately, which are defined as:
\begin{equation}
\begin{split}
\mathcal{L}_{actor}&=-\log \pi_{\theta} (a_t|s_t)(r_t+\gamma v_{t+1}-v_{t})\\
\mathcal{L}_{critic}&=(r_t+\gamma v_{t+1}-v_{t})^2
\end{split}
\end{equation}
where $\gamma$ is a discount factor in our DRL model.
In their definitions, we can find that the actor network aims to maximize the expected rewards to be obtained, and the critic network is minimize the temporal difference error due during the stochastic learning process.

\emph{Remarks:} Deep neural network (DNN) and Deep reinforcement learning (DRL) based algorithms has been widely applied in a variety of applications in many sequential learning tasks, like: system identification \cite{narendra2016fast,wang2017new}, sequence tagging \cite{wang2018deep}, slot filling \cite{wang2018bi}; other NLP tasks like: conference resoluton \cite{clark2016deep}, information extractionand \cite{narasimhan2016improving}, semantic parsing \cite{andreas2016learning} and text games \cite{narasimhan2015language}. There are also many approaches to boost their performance, like: adaptive boosting\cite{ratsch2001soft,hastie2009multi} and multiple models \cite{wang2018boosting,narendra2015improving,narendra2014stability}. In our algorithm, we only use a relative standard actor-critic DRL algorithm as the stress of this paper is on the novelty of its application on a QA problem instead of a new DRL algorithm.
\section{Experiment}
\subsection{Datasets}
We conduct our experiment on four QA datasets which for machine comprehension based QA tasks: WIKIREADING, WIKIREADING LONG, CNN, and SQuAD1.1.

WIKIREADING  \cite{hewlett2016wikireading} is a public question answering dataset generated from Wikidata and Wikipedia. It consolidated all Wikidata statements with the same item and property into a single (item, property, answer) triple, where answer is a set of values. Replacing each item with the text of the linked Wikipedia article (discarding unlinked items) yields a dataset of 18.58M (document, property, answer) instances. 

In order to compare fairly to our baseline models and demonstrate system's performance on long documents, we perform the same operation on the WIKIREADING datasets as in \cite{choi2017coarse} to generate the WIKIREADING LONG dataset. From the WIKIREADING dataset, documents with less than ten sentences are filtered, and only the Wikidata properties for which \cite{hewlett2016wikireading}'s best model obtains an accuracy of less than 60 $\%$ are considered. This prunes out some properties such as GIVEN NAME, INSTANCE OF and GENDER. The resulting WIKIREADING LONG dataset contains 1.97M instances (\emph{i.e.} (document, property, answer) pairs), and only 31$\%$  of the answers are in the first sentence. 

CNN is a question answering dataset containing news stories with associated queries collected from the CNN websites starting from April 2007 till April 2015. There are a total of 90,266 documents associated with more than 380k queries in the training dataset, and 1093 documents with more than 3k associated questions in test dataset. Each document companies on average 4 questions approximately. Each question is a sentence with one missing word/phrase which can be found from the accompanying document/context. It is a commonly used dataset to test the performance of models for machine comprehension based QA tasks.

Stanford Question Answering Dataset (SQuAD) is another well-known reading comprehension dataset, consisting of questions posed by crowdworkers on a set of Wikipedia articles, where the answer to every question is a segment of text, or span, from the corresponding reading passage. SQuAD contrains 107,785 question-answer pairs on 536 articles. Currently, there are two versions of SQuAD datasets: v1.1 and v2.0. The main difference between these two datasets is the SQuAD2.0 added a new category named as "unanswerable questions" to reject the questions which are not in the context and answerable. In order to fairly compare with the baseline models which have never been tested on v2.0 data,  we use the SQuAD1.1 in our experiment.

On average, each document in WIKIREADING contains 490 tokens, the filtered WIKIREADING LONG contains 1.2k tokens, those in CNN contains 763 tokens, and SQuAD only contains 122 tokens. 

Table \ref{table:comparisonDataset} shows a detailed statistics and comparison between these four datasets.

\begin{table}[ht]\scriptsize
\parbox{1\linewidth}{
\centering
	\caption{Statistics of Different QA Datasets}
	\label{table:comparisonDataset}
	\begin{tabular}{>{\centering\arraybackslash}p{1.5cm}|>{\centering\arraybackslash}p{1.2cm}|>{\centering\arraybackslash}p{1cm}|>{\centering\arraybackslash}p{1.5cm}|>{\centering\arraybackslash}p{1.3cm}}
		\toprule
		
		\multirow{2}{*}{\textbf{Datasets}} &\textbf{$\#$ of documents}& \textbf{$\#$ of QA pairs}& \textbf{$\#$ of words/question}&\textbf{$\#$ of tokens/doc.}\\
		
		\midrule
		\midrule
		\multirow{2}{*} {}WIKIREADING & 4.7M &18.58M &2.35&489 \\ 
		
		\multirow{2}{*}{}WR-LONG  &3.9M&1.97M&2.14&1.2K  \\

		\multirow{2}{*}{}CNN &92K & 387K& 12.5 & 763  \\		
		\multirow{2}{*}{}SQuAD  &536 & 107K& 4.53& 122	\\
		\bottomrule		
	\end{tabular}
}
\end{table}
\subsection{Experiment Setup}
For the WIKIREADING \cite{hewlett2016wikireading} and WIKIREADING-LONG \cite{choi2017coarse} datasets, we use the 70$\%$ of the data as the training dataset and 10$\%$ for validation and 20$\%$ for test. For he CNN dataset, we follow the original splits of the train, validation and test set, where 90,266 documents for training, 1,220 documents for validation and 1,093 documents for test. On the SQuAD1.1 \cite{rajpurkar2016squad}\footnote{The SQuAD1.1's test portal is replaced by SQuAD2.0 currently, hence we test our model on the dev dataset in SQuAD1.1 to compare with our baseline models which only have public results on SQuAD1.1}, the results on the dev set is reported for the purpose of comparison with one of our baseline models, \emph{i.e.} QANet, which is specifically designed for the SQuAD1.1 datasetl

The word embedding and character embedding size are 300 and 200 for the input embedding layer to generate the document and question embeddings as in Figure \ref{fig:inputEmbedding}, the output size of the embedding encoder layer is set as 128 .  The filter size of the convolutional neural network in the embedding encoder layer, the model encoder layer are set as 128 and and that in the sentence selection module is set as 100. The CNN kernel size is set as 7 for the embedding encoder layer and the model encoder layer, and set as 5 for the CNN network in sentence selection module.

For the actor-critic model to generate the corresponding actions, we use two Gated Recurrent Units (GRU) neural networks \cite{chung2014empirical} to model the actor and critic separately. Both of the models take the current state $s_t$, \emph{i.e.} the concatenation of current document $D_i$ and question $Q$ as their inputs. The actor GRU generates three action probabilities, and the critic GRU generates an expected reward value based on the given input. The GRU state cells is chosen as 512 for both networks, and the learning rate is 0.0001. The discounted factor in defining loss functions $\mathcal{L}_1$ and $\mathcal{L}_2$ is chosen as $\gamma=0.9$.
AdaDelta \cite{zeiler2012adadelta} optimizer is selected to smooth the convergence of gradient during training.
\subsection{Evaluation Metrics} The main evaluation metric of our experiment is the exact match (EM) accuracy, the proportion of predicted answers match exactly to their ground truth. During the training, if the system choose action $a_1$, our system triggers the terminal state-answer generation module and obtain a predicted answer. If the system choose action $a_2$ and $a_3$, the system will loop back the selected sentences $D_s$ or the sub-context $D_c$, and use it as the document context $D_{i+1}$ for the action selector in next step. To constrain the training and evaluation time, up to $K=5$ action steps is allowed for each question-document pair. If the system still doesn't select the action to trigger the answer generation module ($M_2$) after 5 rounds, we will force the system to call the $M_2$ module and generate an answer based on the extracted document context $D_5$, then evaluate on this answer.
\subsection{Model and Baselines}
Two baseline models are used to compare with our system as shown below:

\underline{Baseline model 1:} The first baseline we choose is a re-implementation of the coarse-to-fine model by \cite{choi2017coarse}. In this work, the author uses the sentence selector as a coarse selector, and generate the answer based on the selected sentences. The sentence selector is the same as our module $M_1$, and the answer generator model is a GRU based sequence to sequence model as described in \cite{choi2017coarse}. Since in our model, we also use sentence selector as one of the three states, it is necessary to compare our model with this model to demonstrate the need of an action selector and other two states. When re-implementing this baseline model, we use the convolutional neural network model as its sentence selection model, and hard attention for the document summary. The predefined number of selected sentences is $K=5$.  

\underline{Baseline model 2:} The second baseline we choose is the QANet given by \cite{wei2018fast}, which also gives strong performance on the SQuAD dataset. Since our answer generator model follows the structure of QANet, it is also necessary to compare our model with this baseline model for all datasets.  When re-implementing this baseline model, we choose the number of convolutional layer in embedding encoder layer as 1 and don't use any data augmentation for training.

\underline{ORACLE:} This model select the sentence containing the answer string (coarse) and generate the answer using QANet.
\subsection{Answer Accuracy Results}
Table \ref{table:accuracyComparison} gives an accuracy comparison between the baseline models and our proposed model. To demonstrate the importance of learning to reject the false-positive answers, we also build another comparative model  without action $a_3$ and the sub-context selection module, named as CFQA (w/o sub-context). All the other model hyper-parameters are the same as CFQA.
\begin{table}[ht]\scriptsize
\parbox{1\linewidth}{
\centering
	\caption{Answer prediction accuracy on the test set (dev set for SQuAD1.1)}
	\label{table:accuracyComparison}
	\begin{tabular}{>{\centering\arraybackslash}p{2cm}|>{\centering\arraybackslash}p{2.3cm}|>{\centering\arraybackslash}p{2.3cm}}
		\toprule
		
		\multirow{2}{*}\textbf{Datasets} &\textbf{Models}& \textbf{Accuracy ($\%$)}\\
		
		\midrule
		\midrule
		\multirow{5}{*} {} & Base Model 1 (K=5) & 73.9  \\ 
		& Base Model 2 &74.1 \\ 
		WIKIREADING & Oracle &74.6  \\ 
		& CFQA &{\bf{75.8}}  \\
		& CFQA (w/o sub-context) &74.3  \\  
			\midrule
		\multirow{5}{*} {} & Base Model 1 (K=5)&42.3  \\ 
		& Base Model 2 &39.5  \\ 
		WR-LONG & Oracle &43.9 \\ 
		& CFQA& {\bf{43.6}} \\
		& CFQA (w/o sub-context) &42.8 \\  

		\midrule
		\multirow{5}{*} {} & Base Model 1 (K=5)&75.8  \\ 
		& Base Model 2 & 72.4 \\ 
		CNN & Oracle &78.8  \\ 
		& CFQA &\bf{77.6}  \\
		& CFQA (w/o sub-context) &76.3  \\  	
		
		\midrule
		\multirow{5}{*} {} & Base Model 1 &75.2 (dev)  \\ 
		& Base Model 2 &82.9 (dev)  \\ 
		SQuAD & Oracle & 86.2 (dev)  \\ 
		& CFQA &\bf{83.6} (dev)  \\
		& CFQA (w/o sub-context) &80.5 (dev)  \\  
		\bottomrule		
	\end{tabular}
}
\end{table}
From Table \ref{table:accuracyComparison}, several experiment observations are obtained:

1. It can be observed that our CFQA performs better than all other baseline models on four QA datasets including both long and short documents. One reason is because that the model can efficiently choose the correct actions to perform by considering the length of the document. The proportions of choosing different actions on four datasets are given in Table \ref{table:pComparison}.

2. The CFQA model performs better than CFQA without sub-context on all datasets, which shows that importance of removing the false-positive answers using another action $a_3$.

3. Base model 1 has relative better performance on long documents ( WR-LONG, CNN) compared to base model 2, and worse performance on shorter documents (WIKIREADING, SQuAD). The CFQA model can handle both cases well by taking the advantages from two baseline models
\subsection{Speed Comparison}
We further compare the training speed with two baseline models on four datasets. The results are shown as in Table \ref{table:speedComparison}.

\begin{table}[ht]\scriptsize
\parbox{1\linewidth}{
\centering
	\caption{Comparison of Training Speed on Different Datasets}
	\label{table:speedComparison}
	\begin{tabular}{>{\centering\arraybackslash}p{2cm}|>{\centering\arraybackslash}p{2.3cm}|>{\centering\arraybackslash}p{2.3cm}}
		\toprule
		
		\multirow{2}{*}\textbf{Datasets} &\textbf{Models}& \textbf{Training Speed(samples/sec)}\\
		
		\midrule
		\midrule
		\multirow{3}{*} {} & Base Model 1 & 112 \\ 
		WIKIREADING& Base Model 2 & 89\\ 
		& CFQA &{\bf{186}}  \\
			\midrule
		\multirow{3}{*} {} & Base Model 1&66 \\ 
		WR-LONG& Base Model 2 &38 \\ 
		& CFQA& {\bf{129}} \\

		\midrule
		\multirow{3}{*} {} & Base Model 1 &83  \\ 
		CNN& Base Model 2 & 67 \\ 
		& CFQA &\bf{173}  \\
		
		\midrule
		\multirow{3}{*} {} & Base Model 1 &146\\ 
		SQuAD & Base Model 2 & 102 \\  
		& CFQA &\bf{213}  \\
		\bottomrule		
	\end{tabular}
}
\end{table}

\begin{table}[ht]\scriptsize
\begin{minipage}{.48\textwidth}
\centering
	\caption{Proportions of Three Actions}
	\label{table:pComparison}
	\begin{tabular}{>{\centering\arraybackslash}p{1.6cm}|>{\centering\arraybackslash}p{1.6cm}|>{\centering\arraybackslash}p{1.6cm}|>{\centering\arraybackslash}p{1.6cm}}
		\toprule
		
		\multirow{2}{*}\textbf{Datasets} &\textbf{Choosing $a_1$ ($\%$)}& \textbf{Choosing $a_2$ ($\%$)}& \textbf{Choosing $a_3$ ($\%$)}\\
		
		\midrule
		\midrule
		\multirow{2}{*} {}WIKIREADING & 67 & 22& 11  \\ 
		
		\multirow{2}{*}{}WR-LONG  &48&39&13   \\

		\multirow{2}{*}{}CNN &55& 26 & 19  \\		
		\multirow{2}{*}{}SQuAD  &76 & 8 &16	\\
		\bottomrule		
	\end{tabular}

\end{minipage}

\begin{minipage}{.48\textwidth}
\hspace{9cm}
\centering
	\caption{Number of Action Steps on Different Datasets}
	\label{table:stepComparison}
	\begin{tabular}{>{\centering\arraybackslash}p{2cm}|>{\centering\arraybackslash}p{2.3cm}|>{\centering\arraybackslash}p{2.3cm}}
		\toprule
		
		\multirow{2}{*}{\textbf{Datasets}} &\textbf{Avg. steps on training dataset ($\leq$ 5)}& \textbf{Avg. Steps on test/ dev dataset}\\
		
		\midrule
		\midrule
		\multirow{2}{*} {}WIKIREADING & 2.9 &3.0   \\ 
		
		\multirow{2}{*}{}WR-LONG  &3.8&3.9   \\

		\multirow{2}{*}{}CNN &3.3& 3.2  \\		
		\multirow{2}{*}{}SQuAD  &1.8 & 1.9 (dev)	\\
		\bottomrule		
	\end{tabular}
\end{minipage}
\end{table}
From Table \ref{table:speedComparison}, it shows that the CFQA model improve the training speed by 1.5x-3.4x compared to the baseline models. It is worth noticing that the baseline models are among the fastest QA models on WIKIREADING and SQuAD datasets. The speed-ups are mainly from the improvement on the model's efficiency by choosing appropriate  actions based on documents' lengths.
\subsection{Evaluation on Action Selector}
To further evaluate how well the DRL based action selector performs in our CFQA model structure, we show the statistics of proportions of three different actions chose by our system during test in Table \ref{table:pComparison} . These numbers are calculated based as: 
\begin{equation}
p_{a_i}=\dfrac{\#\text{ of times choosing action } a_i}{\sum \#\text{ of actions selected at all steps}}
\end{equation}


Based on the statistics given in Table \ref{table:pComparison}, one can conclude that:

1. When the datasets contain more longer documents, the system tend to select action $a_2$ in order to firstly select sentences in a coarse manner.

2. When the datasets contain more shorter documents, the system tend to select action $a_1$ in order to answer the question directly without coarsely select some sentences. 

In order to better understand the DRL action selector's behavior, we use Table \ref{table:stepComparison} to show the average number of steps performed during training and test/dev on each dataset.

It can be observed that, datatsets containing more long documents with higher average tokens per document tend to have more average action steps. By combining Table \ref{table:pComparison} and \ref{table:stepComparison}, it is shown that the multi-step DRL structure with multiple actions is useful for our model to choose a more suitable module to perform, in order to get a better performance.
\section{Conclusion}
In this paper, we present a multi-step coarse to fine question answering (CFQA) system which can efficiently process both long and short documents by choosing appropriate actions. The system shows decent results on four different QA datasets in terms of accuracy and training speed. It also gives a new concept of using DRL model to guide a multi-step QA reasoning process, which is more close to human-being's judgment behavior. In the future, we would like to investigate more on refining the design of system by adding more possible actions and states, such that the system can behave even smarter and faster.

\bibliographystyle{aaai}
\bibliography{aaai19MM}

\begin{thebibliography}{}

\bibitem[\protect\citeauthoryear{Andreas \bgroup et al\mbox.\egroup
  }{2016}]{andreas2016learning}
Andreas, J.; Rohrbach, M.; Darrell, T.; and Klein, D.
\newblock 2016.
\newblock Learning to compose neural networks for question answering.
\newblock {\em arXiv preprint arXiv:1601.01705}.

\bibitem[\protect\citeauthoryear{Chen \bgroup et al\mbox.\egroup
  }{2017}]{chen2017reading}
Chen, D.; Fisch, A.; Weston, J.; and Bordes, A.
\newblock 2017.
\newblock Reading wikipedia to answer open-domain questions.
\newblock In {\em Proceedings of the 55th Annual Meeting of the Association for
  Computational Linguistics (Volume 1: Long Papers)}, volume~1,  1870--1879.

\bibitem[\protect\citeauthoryear{Chen, Bolton, and
  Manning}{2016}]{chen2016thorough}
Chen, D.; Bolton, J.; and Manning, C.~D.
\newblock 2016.
\newblock A thorough examination of the cnn/daily mail reading comprehension
  task.
\newblock {\em arXiv preprint arXiv:1606.02858}.

\bibitem[\protect\citeauthoryear{Choi \bgroup et al\mbox.\egroup
  }{2017}]{choi2017coarse}
Choi, E.; Hewlett, D.; Uszkoreit, J.; Polosukhin, I.; Lacoste, A.; and Berant,
  J.
\newblock 2017.
\newblock Coarse-to-fine question answering for long documents.
\newblock In {\em Proceedings of the 55th Annual Meeting of the Association for
  Computational Linguistics (Volume 1: Long Papers)}, volume~1,  209--220.

\bibitem[\protect\citeauthoryear{Chung \bgroup et al\mbox.\egroup
  }{2014}]{chung2014empirical}
Chung, J.; Gulcehre, C.; Cho, K.; and Bengio, Y.
\newblock 2014.
\newblock Empirical evaluation of gated recurrent neural networks on sequence
  modeling.
\newblock {\em arXiv preprint arXiv:1412.3555}.

\bibitem[\protect\citeauthoryear{Clark and Manning}{2016}]{clark2016deep}
Clark, K., and Manning, C.~D.
\newblock 2016.
\newblock Deep reinforcement learning for mention-ranking coreference models.
\newblock {\em arXiv preprint arXiv:1609.08667}.

\bibitem[\protect\citeauthoryear{Hastie \bgroup et al\mbox.\egroup
  }{2009}]{hastie2009multi}
Hastie, T.; Rosset, S.; Zhu, J.; and Zou, H.
\newblock 2009.
\newblock Multi-class adaboost.
\newblock {\em Statistics and its Interface} 2(3):349--360.

\bibitem[\protect\citeauthoryear{Hermann \bgroup et al\mbox.\egroup
  }{2015}]{hermann2015teaching}
Hermann, K.~M.; Kocisky, T.; Grefenstette, E.; Espeholt, L.; Kay, W.; Suleyman,
  M.; and Blunsom, P.
\newblock 2015.
\newblock Teaching machines to read and comprehend.
\newblock In {\em Advances in Neural Information Processing Systems},
  1693--1701.

\bibitem[\protect\citeauthoryear{Hewlett \bgroup et al\mbox.\egroup
  }{2016}]{hewlett2016wikireading}
Hewlett, D.; Lacoste, A.; Jones, L.; Polosukhin, I.; Fandrianto, A.; Han, J.;
  Kelcey, M.; and Berthelot, D.
\newblock 2016.
\newblock Wikireading: A novel large-scale language understanding task over
  wikipedia.
\newblock In {\em Proceedings of the 54th Annual Meeting of the Association for
  Computational Linguistics (Volume 1: Long Papers)}, volume~1,  1535--1545.

\bibitem[\protect\citeauthoryear{Hill \bgroup et al\mbox.\egroup
  }{2015}]{hill2015goldilocks}
Hill, F.; Bordes, A.; Chopra, S.; and Weston, J.
\newblock 2015.
\newblock The goldilocks principle: Reading children's books with explicit
  memory representations.
\newblock {\em arXiv preprint arXiv:1511.02301}.

\bibitem[\protect\citeauthoryear{Kadlec \bgroup et al\mbox.\egroup
  }{2016}]{kadlec2016text}
Kadlec, R.; Schmid, M.; Bajgar, O.; and Kleindienst, J.
\newblock 2016.
\newblock Text understanding with the attention sum reader network.
\newblock In {\em Proceedings of the 54th Annual Meeting of the Association for
  Computational Linguistics (Volume 1: Long Papers)}, volume~1,  908--918.

\bibitem[\protect\citeauthoryear{Konda and Tsitsiklis}{2000}]{konda2000actor}
Konda, V.~R., and Tsitsiklis, J.~N.
\newblock 2000.
\newblock Actor-critic algorithms.
\newblock In {\em Advances in neural information processing systems},
  1008--1014.

\bibitem[\protect\citeauthoryear{Miller \bgroup et al\mbox.\egroup
  }{2016}]{miller2016key}
Miller, A.; Fisch, A.; Dodge, J.; Karimi, A.-H.; Bordes, A.; and Weston, J.
\newblock 2016.
\newblock Key-value memory networks for directly reading documents.
\newblock {\em arXiv preprint arXiv:1606.03126}.

\bibitem[\protect\citeauthoryear{Mnih \bgroup et al\mbox.\egroup
  }{2015}]{mnih2015human}
Mnih, V.; Kavukcuoglu, K.; Silver, D.; Rusu, A.~A.; Veness, J.; Bellemare,
  M.~G.; Graves, A.; Riedmiller, M.; Fidjeland, A.~K.; Ostrovski, G.; et~al.
\newblock 2015.
\newblock Human-level control through deep reinforcement learning.
\newblock {\em Nature} 518(7540):529.

\bibitem[\protect\citeauthoryear{Narasimhan, Kulkarni, and
  Barzilay}{2015}]{narasimhan2015language}
Narasimhan, K.; Kulkarni, T.; and Barzilay, R.
\newblock 2015.
\newblock Language understanding for text-based games using deep reinforcement
  learning.
\newblock {\em arXiv preprint arXiv:1506.08941}.

\bibitem[\protect\citeauthoryear{Narasimhan, Yala, and
  Barzilay}{2016}]{narasimhan2016improving}
Narasimhan, K.; Yala, A.; and Barzilay, R.
\newblock 2016.
\newblock Improving information extraction by acquiring external evidence with
  reinforcement learning.
\newblock {\em arXiv preprint arXiv:1603.07954}.

\bibitem[\protect\citeauthoryear{Narendra, Mukhopadyhay, and
  Wang}{2015}]{narendra2015improving}
Narendra, K.~S.; Mukhopadyhay, S.; and Wang, Y.
\newblock 2015.
\newblock Improving the speed of response of learning algorithms using multiple
  models.
\newblock {\em arXiv preprint arXiv:1510.05034}.

\bibitem[\protect\citeauthoryear{Narendra, Wang, and
  Chen}{2014}]{narendra2014stability}
Narendra, K.~S.; Wang, Y.; and Chen, W.
\newblock 2014.
\newblock Stability, robustness, and performance issues in second level
  adaptation.
\newblock In {\em American Control Conference (ACC), 2014},  2377--2382.
\newblock IEEE.

\bibitem[\protect\citeauthoryear{Narendra, Wang, and
  Mukhopadhay}{2016}]{narendra2016fast}
Narendra, K.~S.; Wang, Y.; and Mukhopadhay, S.
\newblock 2016.
\newblock Fast reinforcement learning using multiple models.
\newblock In {\em Decision and Control (CDC), 2016 IEEE 55th Conference on},
  7183--7188.
\newblock IEEE.

\bibitem[\protect\citeauthoryear{Nguyen \bgroup et al\mbox.\egroup
  }{2016}]{nguyen2016ms}
Nguyen, T.; Rosenberg, M.; Song, X.; Gao, J.; Tiwary, S.; Majumder, R.; and
  Deng, L.
\newblock 2016.
\newblock Ms marco: A human generated machine reading comprehension dataset.
\newblock {\em arXiv preprint arXiv:1611.09268}.

\bibitem[\protect\citeauthoryear{Onishi \bgroup et al\mbox.\egroup
  }{2016}]{onishi2016did}
Onishi, T.; Wang, H.; Bansal, M.; Gimpel, K.; and McAllester, D.
\newblock 2016.
\newblock Who did what: A large-scale person-centered cloze dataset.
\newblock {\em arXiv preprint arXiv:1608.05457}.

\bibitem[\protect\citeauthoryear{Pennington, Socher, and
  Manning}{2014}]{pennington2014glove}
Pennington, J.; Socher, R.; and Manning, C.
\newblock 2014.
\newblock Glove: Global vectors for word representation.
\newblock In {\em Proceedings of the 2014 conference on empirical methods in
  natural language processing (EMNLP)},  1532--1543.

\bibitem[\protect\citeauthoryear{Peters and Schaal}{2008}]{peters2008natural}
Peters, J., and Schaal, S.
\newblock 2008.
\newblock Natural actor-critic.
\newblock {\em Neurocomputing} 71(7-9):1180--1190.

\bibitem[\protect\citeauthoryear{Rajpurkar \bgroup et al\mbox.\egroup
  }{2016}]{rajpurkar2016squad}
Rajpurkar, P.; Zhang, J.; Lopyrev, K.; and Liang, P.
\newblock 2016.
\newblock Squad: 100,000+ questions for machine comprehension of text.
\newblock {\em arXiv preprint arXiv:1606.05250}.

\bibitem[\protect\citeauthoryear{R{\"a}tsch, Onoda, and
  M{\"u}ller}{2001}]{ratsch2001soft}
R{\"a}tsch, G.; Onoda, T.; and M{\"u}ller, K.-R.
\newblock 2001.
\newblock Soft margins for adaboost.
\newblock {\em Machine learning} 42(3):287--320.

\bibitem[\protect\citeauthoryear{Seo \bgroup et al\mbox.\egroup
  }{2016}]{seo2016bidirectional}
Seo, M.; Kembhavi, A.; Farhadi, A.; and Hajishirzi, H.
\newblock 2016.
\newblock Bidirectional attention flow for machine comprehension.
\newblock {\em arXiv preprint arXiv:1611.01603}.

\bibitem[\protect\citeauthoryear{Sutton \bgroup et al\mbox.\egroup
  }{2000}]{sutton2000policy}
Sutton, R.~S.; McAllester, D.~A.; Singh, S.~P.; and Mansour, Y.
\newblock 2000.
\newblock Policy gradient methods for reinforcement learning with function
  approximation.
\newblock In {\em Advances in neural information processing systems},
  1057--1063.

\bibitem[\protect\citeauthoryear{Trischler \bgroup et al\mbox.\egroup
  }{2016}]{trischler2016newsqa}
Trischler, A.; Wang, T.; Yuan, X.; Harris, J.; Sordoni, A.; Bachman, P.; and
  Suleman, K.
\newblock 2016.
\newblock Newsqa: A machine comprehension dataset.
\newblock {\em arXiv preprint arXiv:1611.09830}.

\bibitem[\protect\citeauthoryear{Vaswani \bgroup et al\mbox.\egroup
  }{2017}]{vaswani2017attention}
Vaswani, A.; Shazeer, N.; Parmar, N.; Uszkoreit, J.; Jones, L.; Gomez, A.~N.;
  Kaiser, {\L}.; and Polosukhin, I.
\newblock 2017.
\newblock Attention is all you need.
\newblock In {\em Advances in Neural Information Processing Systems},
  5998--6008.

\bibitem[\protect\citeauthoryear{Wang and Jin}{2018}]{wang2018boosting}
Wang, Y., and Jin, H.
\newblock 2018.
\newblock A boosting-based deep neural networks algorithm for reinforcement
  learning.
\newblock In {\em American Control Conference (ACC)}.

\bibitem[\protect\citeauthoryear{Wang \bgroup et al\mbox.\egroup
  }{2018}]{wang2018deep}
Wang, Y.; Patel, A.; Shen, Y.; and Jin, H.
\newblock 2018.
\newblock A deep reinforcement learning based multimodal coaching model (dcm)
  for slot filling in spoken language understanding (slu).
\newblock {\em Proc. Interspeech 2018}  3444--3448.

\bibitem[\protect\citeauthoryear{Wang, Shen, and Jin}{2018}]{wang2018bi}
Wang, Y.; Shen, Y.; and Jin, H.
\newblock 2018.
\newblock A bi-model based rnn semantic frame parsing model for intent
  detection and slot filling.
\newblock In {\em Proceedings of the 2018 Conference of the North American
  Chapter of the Association for Computational Linguistics: Human Language
  Technologies, Volume 2 (Short Papers)}, volume~2,  309--314.

\bibitem[\protect\citeauthoryear{Wang}{2017}]{wang2017new}
Wang, Y.
\newblock 2017.
\newblock A new concept using lstm neural networks for dynamic system
  identification.
\newblock In {\em American Control Conference (ACC)},  5324--5329.

\bibitem[\protect\citeauthoryear{Weissenborn, Wiese, and
  Seiffe}{2017}]{weissenborn2017making}
Weissenborn, D.; Wiese, G.; and Seiffe, L.
\newblock 2017.
\newblock Making neural qa as simple as possible but not simpler.
\newblock In {\em Proceedings of the 21st Conference on Computational Natural
  Language Learning (CoNLL 2017)},  271--280.

\bibitem[\protect\citeauthoryear{Xiong, Zhong, and
  Socher}{2016}]{xiong2016dynamic}
Xiong, C.; Zhong, V.; and Socher, R.
\newblock 2016.
\newblock Dynamic coattention networks for question answering.
\newblock {\em arXiv preprint arXiv:1611.01604}.

\bibitem[\protect\citeauthoryear{Yang \bgroup et al\mbox.\egroup
  }{2016}]{yang2016hierarchical}
Yang, Z.; Yang, D.; Dyer, C.; He, X.; Smola, A.; and Hovy, E.
\newblock 2016.
\newblock Hierarchical attention networks for document classification.
\newblock In {\em Proceedings of the 2016 Conference of the North American
  Chapter of the Association for Computational Linguistics: Human Language
  Technologies},  1480--1489.

\bibitem[\protect\citeauthoryear{Yu \bgroup et al\mbox.\egroup
  }{2014}]{yu2014deep}
Yu, L.; Hermann, K.~M.; Blunsom12, P.; and Pulman, S.
\newblock 2014.
\newblock Deep learning for answer sentence selection.
\newblock {\em arXiv preprint arXiv:1412.1632}.

\bibitem[\protect\citeauthoryear{Yu \bgroup et al\mbox.\egroup
  }{2018}]{wei2018fast}
Yu, A.~W.; Dohan, D.; Le, Q.; Luong, T.; Zhao, R.; and Chen, K.
\newblock 2018.
\newblock Fast and accurate reading comprehension by combining self-attention
  and convolution.
\newblock In {\em International Conference on Learning Representations}.

\bibitem[\protect\citeauthoryear{Zeiler}{2012}]{zeiler2012adadelta}
Zeiler, M.~D.
\newblock 2012.
\newblock Adadelta: an adaptive learning rate method.
\newblock {\em arXiv preprint arXiv:1212.5701}.

\end{thebibliography}

\end{document}